\documentclass{article} 
\usepackage[preprint]{colm2026_conference}

\usepackage{microtype}
\usepackage{hyperref}
\usepackage{url}
\usepackage{booktabs}
\usepackage{graphicx}
\usepackage{amsmath,amssymb,amsthm,mathabx,amsfonts}
\usepackage{tcolorbox}
\usepackage{multirow}    
\usepackage[table]{xcolor} 
\usepackage{bbm}
\usepackage{arydshln}
\usepackage{enumitem}
\usepackage{pifont}
\usepackage{xspace}
\usepackage{wrapfig}
\usepackage{caption}

\usepackage{lineno}

\definecolor{darkblue}{rgb}{0, 0, 0.5}
\hypersetup{colorlinks=true, citecolor=darkblue, linkcolor=darkblue, urlcolor=darkblue}

\definecolor{myyellow}{rgb}{.99,.94,.82}
\definecolor{tabcolor1}{RGB}{247,225, 237} 
\definecolor{tabcolor2}{RGB}{255, 250, 132} 
\definecolor{tabcolor3}{RGB}{204, 232, 207} 
\definecolor{tabcolor4}{RGB}{245, 222, 179} 
\definecolor{tabcolor5}{RGB}{210, 220, 250} 
\definecolor{tabcolor6}{RGB}{222, 222, 222} 
\newcommand{\tightbox}[2]{%
    \setlength{\fboxsep}{1pt}%
    \smash{%
        \colorbox{#1}{#2}%
    }%
    \vphantom{#2}%
}

\title{KARL: Mitigating Hallucinations in LLMs via \\ \underline{K}nowledge-Boundary-\underline{A}ware \underline{R}einforcement \underline{L}earning}

\author{Cheng Gao, Cheng Huang, Kangyang Luo, Ziqing Qiao,  \\ \textbf{Shuzheng Si, Huimin Chen, Chaojun Xiao \& Maosong Sun}   \\
Tsinghua University\\
 \texttt{\{gaoc24\}@mails.tsinghua.edu.cn}, \texttt{\{huimchen,xcj,sms\}@tsinghua.edu.cn} \\
}
\newcommand{\rewardname}{\textsc{KAR}}
\newcommand{\framename}{\textsc{KARL}}

\begin{document}

\ifcolmsubmission
\linenumbers
\fi

\maketitle

\begin{abstract}
Enabling large language models (LLMs) to appropriately abstain from answering questions beyond their knowledge is crucial for mitigating hallucinations.
While existing reinforcement learning methods foster autonomous abstention, they often compromise answer accuracy because their static reward mechanisms, agnostic to models' knowledge boundaries, drive models toward excessive caution.
In this work, we propose \framename{}, a novel framework that continuously aligns an LLM's abstention behavior with its evolving knowledge boundary. \framename{} introduces two core innovations: a \textbf{\textit{Knowledge-Boundary-Aware Reward}} that performs online knowledge boundary estimation using within-group response statistics, dynamically rewarding correct answers or guided abstention; and a \textbf{\textit{Two-Stage RL Training Strategy}} that first explores the knowledge boundary and bypasses the ``\textit{abstention trap}'', and subsequently converts incorrect answers beyond the knowledge boundary into abstentions without sacrificing accuracy.
Extensive experiments on multiple benchmarks demonstrate that \framename{} achieves a superior accuracy-hallucination trade-off, effectively suppressing hallucinations while maintaining high accuracy across both in-distribution and out-of-distribution scenarios.
\end{abstract}

\begin{wrapfigure}{r}{0.49\textwidth}
  \centering
  \vspace*{-1ex}
  \includegraphics[width=0.48\textwidth]{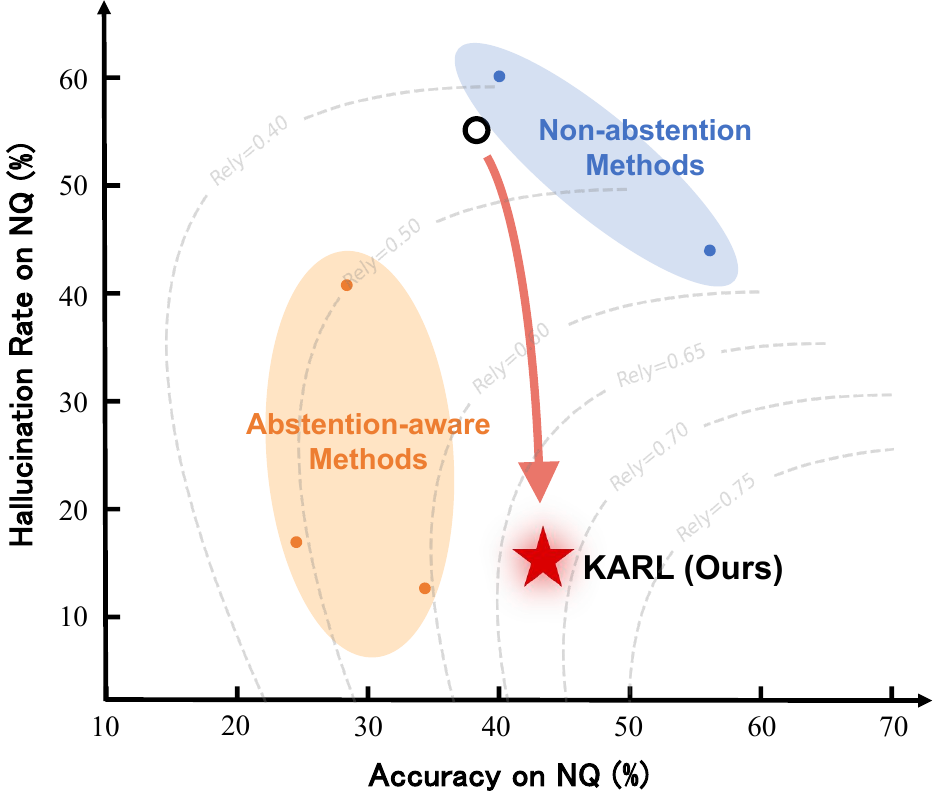}
  \caption{
        Hallucination rate vs. accuracy on NaturalQuestions (NQ)~\citep{NQ}. \textbf{\framename{}} achieves a better trade-off by increasing accuracy while reducing hallucination compared to existing non-abstention and abstention-aware methods.
        Dashed isocontours represent constant Rely scores. 
    }
  \label{fig:fig1}
  \vspace*{-3ex}
\end{wrapfigure}

\vspace*{-2ex}

\section{Introduction}
Large language models (LLMs) have demonstrated remarkable capabilities in open-domain, knowledge-intensive tasks~\citep{openai_gpt5_2025,deepmind_gemini3pro_2025,Deepseek-r1}. However, their reliability is often undermined by hallucination---the tendency to generate plausible but factually incorrect responses with confidence, rather than admitting they do not know the answer~\citep{OnFaithfulnessandFactuality,hallusurvey}. A key cause of this issue is that conventional training typically penalizes abstention as an incorrect prediction, encouraging models to guess blindly in order to maximize training accuracy even when they lack sufficient knowledge~\citep{OpenaiHallucination}. Such behavior poses serious risks in high-stakes domains like medicine and law, where minor factual errors can lead to harmful consequences~\citep{lawformer,Large_Legal_Fictions,Hallucination-Free,Medical}. Therefore, enabling LLMs to abstain appropriately has emerged as a fundamental objective in safety alignment~\citep{wen2024Knowyourlimits}. 

Recently, Reinforcement Learning (RL) has gained significant attention for mitigating hallucination.
Instead of forcing models to mimic fixed abstention responses, RL methods guide them to discover optimal policies through feedback signals~\citep{DeepReinforcementLearningfromHumanPreferences,RLHF,DPO}.
For example, early efforts employed PPO or DPO to align models with abstention preferences~\citep{HH-RLHF,LLMsKnowWhatTheyKnow,Rejection_Improves_Reliability}.
More recently, Reinforcement Learning with Verifiable Rewards (RLVR) has become a dominant paradigm due to its effectiveness and data efficiency~\citep{DeepSeekMath,DeepSeek-V3,Deepseek-r1}. Building upon this, 
notable works such as TruthRL leverage GRPO~\citep{DeepSeekMath} with a ternary reward to jointly optimize correct answers, abstentions, and incorrect responses~\citep{truthrl}. 
While these methods allow models to autonomously learn when to abstain, they still suffer from substantial degradation in answer accuracy.
The underlying issue is that existing methods often rely on static reward structures that are agnostic to the model's knowledge boundary, which drives policies to converge rapidly toward abstention. As a result, models tend to become overly cautious, suppressing both hallucinations and correct answers. Thus, a central question arises: \textit{\textbf{how to design a training framework that reduces hallucinations while maintaining answer accuracy, i.e., ensuring LLMs abstain only when exceeding their knowledge boundary?}}

To tackle the above issues, we propose \textbf{\framename{}}, a framework that continuously aligns the model's abstention behavior with its evolving knowledge boundary.
\framename{} rests on two key innovations. 
First, we devise a \textbf{\textit{Knowledge-Boundary-Aware Reward} (\rewardname{})} mechanism that uses an online knowledge boundary estimator to assess the model’s competence on each query. While this estimator can be implemented in different ways, we adopt a simple and efficient sampling-based approach that dynamically adjusts the reward strength based on whether correct answers appear in each sampled group, thereby motivating the model to confidently explore questions within its competence while steering it toward abstention when knowledge is genuinely lacking.
Second, to structurally decouple knowledge boundary exploration from abstention calibration, we design a \textbf{\textit{two-stage RL training strategy}}. An initial \textit{knowledge boundary exploration stage} employs a mixed-reward strategy that maximizes accuracy and bypasses the ``\textit{abstention trap}'' (See Fig.~\ref{fig:binary_curve}); a subsequent \textit{abstention calibration stage} then converts persistent errors into appropriate abstentions without erasing prior accuracy gains. Together, these components enable LLMs to abstain only when exceeding their knowledge boundaries, thereby reducing hallucinations while maintaining high accuracy.

Extensive experiments on multiple benchmarks demonstrate that \framename{} achieves a superior accuracy-hallucination trade-off (Figure~\ref{fig:fig1}).
Unlike prior methods that often sacrifice accuracy, \framename{} significantly slashes hallucination while maintaining accuracy, even achieving improvements on certain benchmarks. For instance, reducing hallucinations by 38.6\% on NQ while improving accuracy by 4.0\%.
This capability to suppress hallucinations while maintaining high answer accuracy generalizes robustly across out-of-distribution inputs.

\section{Pilot Study}  
\label{sec:pilot}

Models often struggle to balance factual accuracy with appropriate abstention. In this section, we first formalize the abstention task and reward structures. We then investigate the underlying trade-off between answering and abstaining, focusing on two core aspects:
First, we assess whether binary rewards improve pass@1 in short-form open-domain QA, extending prior successes in reasoning tasks~\citep{NIPS-RL,zhang2025interplaypretrainingmidtrainingrl}. 
Second, we identify a critical failure mode of static ternary rewards: excessive early-stage abstention that leads to irreversible policy collapse.
By tracing the evolution of response distributions, we uncover the mechanistic drift into this ``\textit{abstention trap}''.

\begin{figure*}[t]
  \centering
  \includegraphics[width=1.0\linewidth]{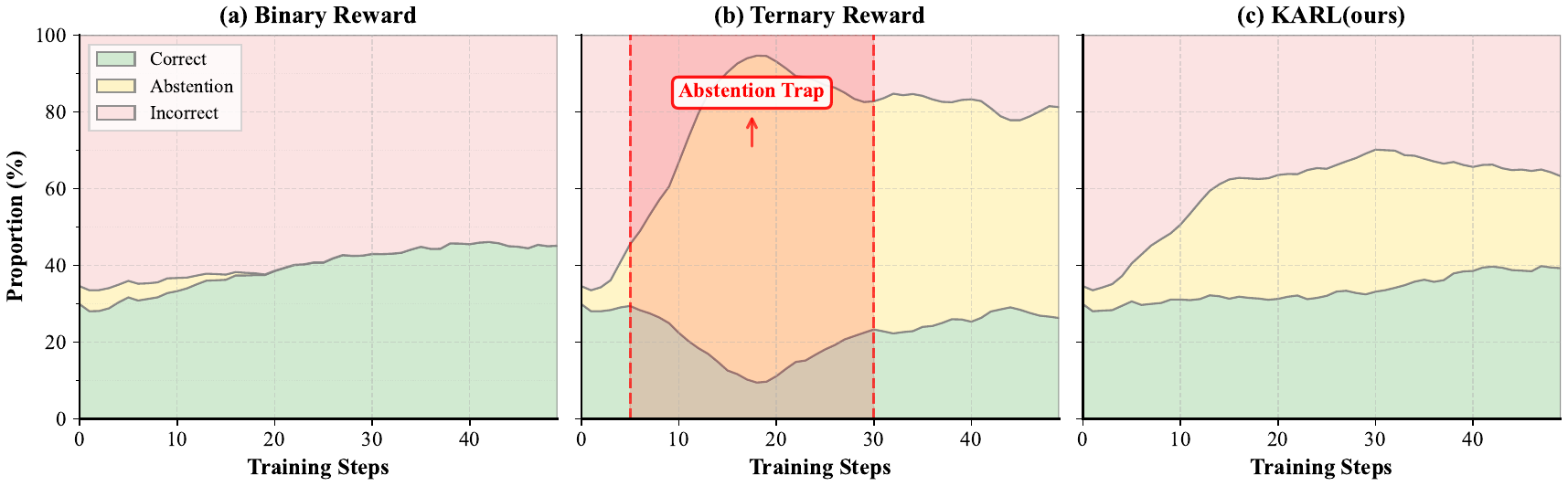}
    \caption{
    Training dynamics of response distributions for Llama-3.1-8B-Instruct on NQ under different reward schemes.
    The plots show the proportions of \textbf{Correct} (green), \textbf{Abstention} (yellow), and \textbf{Incorrect} (red) responses during early training for
    (a) Binary Reward, 
    (b) Static Ternary Reward, and 
    (c) our method. 
    The red dashed region in (b) highlights a rapid increase in abstention accompanied by declining accuracy.
    }
  \label{fig:binary_curve}
\end{figure*}

\paragraph{Task Formulation} Given a factual query $q$, a language model $\pi$ must decide whether to generate an answer $a$ or to abstain from answering (e.g., output ``I don't know''). The goal is to train a model that abstains only when it is likely to produce an incorrect answer, thereby improving reliability while preserving answer accuracy.

Formally, we consider a policy model $\pi_\theta$ which, given $q$, generates a response $y$. The response space is $\mathcal{Y} = \mathcal{A} \cup \{\varnothing\}$, where $\mathcal{A}$ is the set of possible answer texts and $\varnothing$ denotes abstention. Each query $q$ is associated with a ground-truth answer $a^\star$, and the outcome of a response is evaluated with a ternary reward $r$ based on its factual correctness:
\begin{align}
    r(y, a^\star) =
\begin{cases}
r_{\text{pos}}, & \text{if } y \text{ is correct}, \\
r_{\text{neg}}, & \text{if } y \text{ is incorrect}, \\
r_{\text{abs}}, & \text{if } y = \varnothing,
\end{cases}
\end{align}
where $r_{\text{pos}} > r_{\text{abs}} \ge r_{\text{neg}}$. The objective is to optimize the policy $\pi_\theta(y|q)$ to maximize the expected reward $\mathbb{E}_{q, y \sim \pi_\theta}[r(y, a^\star)]$, thus learning when to answer and when to abstain.

\paragraph{Binary Rewards for Knowledge Boundary Exploration.}
Prior work shows RLVR bolsters model confidence in correct reasoning paths, improving pass@1~\citep{NIPS-RL,zhang2025interplaypretrainingmidtrainingrl}. 
To investigate whether this benefit extends to short-form open-domain QA, we adopt GRPO~\citep{DeepSeekMath} as our optimization backbone (formalized in Appendix \ref{app:GRPO}) and train models on NQ with a binary reward (1 for correct answers, 0 otherwise; see Appendix \ref{app:pilot_details} for more details).
As shown in Figure~\ref{fig:binary_curve}(a), abstention plunges to zero early, while accuracy rises steadily thereafter.
Gains persist on out-of-distribution tasks (Table~\ref{tab:main_results}), confirming that binary rewards encourage exploration of knowledge boundary and improve accuracy, but entirely suppress abstention, inducing high hallucination rates.


\begin{wrapfigure}{r}{0.5\textwidth}
  \centering
  \vspace*{-2.8ex}
  \includegraphics[width=0.48\textwidth]{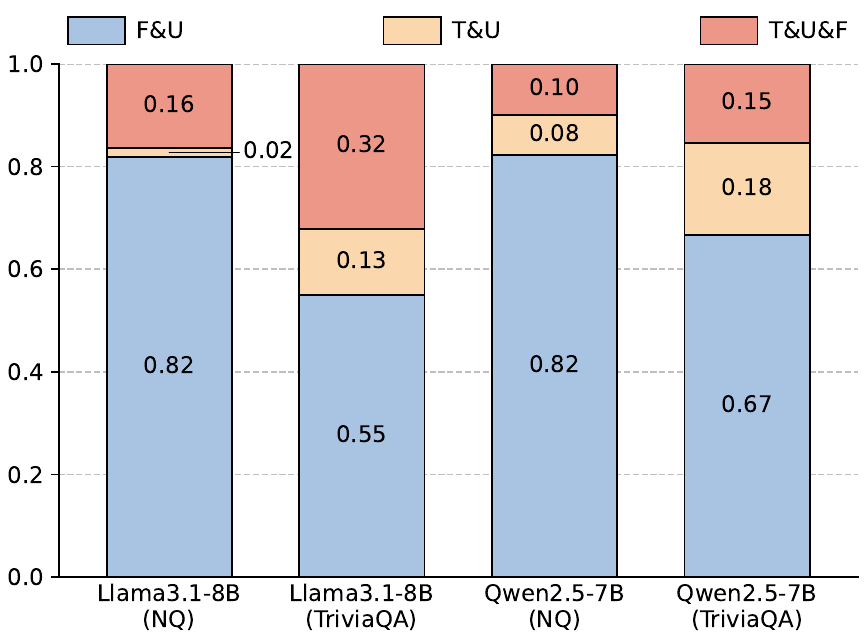}
  \caption{
        Distribution of rollout groups across different models and datasets. 
        We categorize sampled groups based on the composition of Correct (\textbf{T}), Incorrect (\textbf{F}), and Abstention (\textbf{U}) responses. 
    }
  \label{fig:rollout_distribution}
  \vspace*{-2.8ex}
\end{wrapfigure}

\paragraph{Abstention Collapse under Static Ternary Rewards.}
Static ternary rewards (e.g., $\{r_{\text{pos}}=1, r_{\text{abs}}=0, r_{\text{neg}}=-1\}$) are used to induce calibrated abstention~\citep{truthrl}. 
However, they often trigger a critical failure mode: models rapidly maximize abstention at the expense of accuracy, often  abstaining on all queries (Fig.~\ref{fig:binary_curve}(b)). Once this occurs, within-group variance vanishes, halting further learning.

To dissect this collapse, we analyze rollout distributions from Llama-3.1-8B-Instruct~\citep{Llama3} and Qwen2.5-7B-Instruct~\citep{Qwen2.5} on NQ and TriviaQA~\citep{Triviaqa} (see Appendix~\ref{app:pilot_details} for more details). 
We categorize each sampled group based on its composition of Correct (T), Incorrect (F), and Abstention (U) responses. As illustrated in Fig.~\ref{fig:rollout_distribution}, we focus on the relative proportions of heterogeneous groups (F\&U, T\&U, and T\&U\&F), specifically excluding homogeneous groups and T\&F-only groups. We find that groups containing only incorrect and abstention (i.e., F\&U) responses dominate the distribution that influences abstention learning.
Crucially, under group-normalized advantage, any abstention reward $r_{\text{abs}} > r_{\text{neg}}$ yields a consistent positive advantage for abstention in these groups. This creates an overwhelming structural bias early in training, driving the policy into an irreversible ``\textit{abstention trap}'' before it can explore correct answers. We provide further empirical justification in Section~\ref{sec:trap}.

\paragraph{Insights.}  
Our findings confirm that RLVR exploration holds considerable potential for improving factual accuracy. 
Yet, static ternary rewards fail to harness this potential. 
Notably, early in training, models are driven by the predominance of incorrect-and-abstention (F\&U) groups into an ``\textit{abstention trap}'', which prevents accuracy gains and can even degrade performance. 
Thus, a method is needed that both leverages RLVR to enhance accuracy and counteracts the bias from F\&U groups to avoid abstention collapse.


\section{Methodology}

\begin{figure*}[t]
  \centering
  \includegraphics[width=1.0\linewidth]{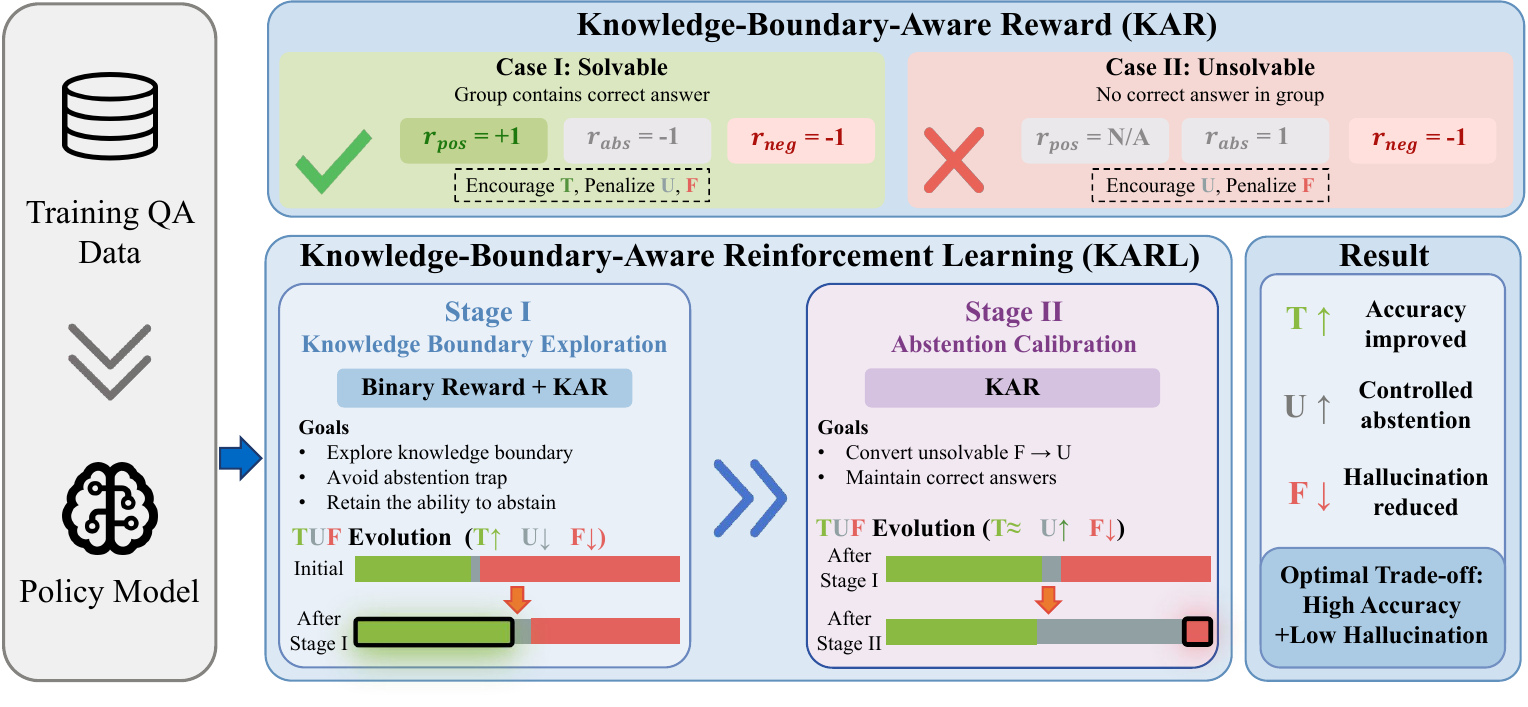}
    \caption{
    An overview of the \framename{} framework.~\framename{} utilizes a Knowledge-Boundary-Aware Reward (\rewardname{}) and employs a two-stage strategy to first explore knowledge boundary and subsequently calibrate abstention behavior.
    }
  \label{fig:framework}
\end{figure*}

In this section, we detail our proposed \framename{} methods, which address the core challenge of aligning a model's abstention behavior with its dynamically evolving knowledge boundary.
As shown in Figure~\ref{fig:framework}, \framename{} consists of two components: (1) \textit{a Knowledge-Boundary-Aware Reward mechanism} designed to provide nuanced learning signals, and (2) \textit{a Two-Stage RL Training Strategy} that decouples knowledge exploration from abstention calibration and prevents premature policy collapse. 

\subsection{Knowledge-Boundary-Aware Reward}
\label{sec:method_reward}
A challenge in abstention alignment is the estimation of whether a given query lies within the model's current knowledge boundary.
To estimate this boundary, a well-established method is to sample the model multiple times to check for the presence of at least one correct response; if the model can generate a correct answer even once across several trials, the query is deemed to be within the reach of its potential capability \citep{R-Tuning, KLCF, truthrl}.
However, prior methods typically perform this probing offline on a static training set to pre-label queries as answerable or not. 
Although intuitive, this static approach fails to reflect the model's continuously evolving capability during the alignment process itself. 
To address this, we propose the Knowledge-Boundary-Aware Reward (\textbf{\rewardname{}}), which performs dynamic, online estimation of the boundary by directly leveraging the sampled responses during reinforcement learning. 


Specifically, for each query $q$, the policy generates $G$ responses $\{o_1, \cdots, o_G\}$ before the update, i.e., $\pi_{\theta_{\text{old}}}$.
\rewardname{} treats the ground-truth $a^\star$ to $q$ as a supervisory signal of the model's current capability.
First, we define a binary solvability indicator:
\begin{equation}
		\begin{aligned}
            \mathbb{I}_{\text{solvable}} = 
                \begin{cases} 
                1 & \text{if}~\exists o \in \{o_i\}_{i=1}^G, \text{ such that } o \text{ is correct}, \\
                0 & \text{otherwise}.
                \end{cases}
		\end{aligned}
\end{equation}

Based on this indicator, \rewardname{} provides distinct learning signals to guide the policy toward either confident answering or appropriate abstention. 
For query deemed \textit{solvable} ($\mathbb{I}_{\text{solvable}} = 1$), the goal is to reinforce correct answering and discourage unnecessary abstention; thus, correct answer receives a positive reward ($r_{\text{pos}}=+1.0$), while both incorrect answer and abstention are penalized ($r_{\text{neg}}=r_{\text{abs}}=-1.0$). For query classified as \textit{unknown} ($\mathbb{I}_{\text{solvable}} = 0$), the objective shifts toward encouraging honesty; abstention is rewarded ($r_{\text{abs}}=+1.0$), incorrect answers are penalized ($r_{\text{neg}}=-1.0$), and the correct answer case is irrelevant (N/A) as it does not occur under this condition. Formally, the complete reward $r_{\text{\rewardname{}}}(o, a^\star)$ is defined as: 


\begin{equation}
r_{\text{KAR}}(o, a^\star) = 
\begin{cases} 
(r_{\text{pos}}, r_{\text{abs}}, r_{\text{neg}}) = (+1, -1, -1), & \text{if } \mathbb{I}_{\text{solvable}} = 1, \\
(r_{\text{pos}}, r_{\text{abs}}, r_{\text{neg}}) = (\text{N/A}, +1, -1), & \text{if } \mathbb{I}_{\text{solvable}} = 0.
\end{cases}
\end{equation}

This adaptive reward scheme is seamlessly integrated into the advantage calculation, providing a dynamic, within-group learning signal that continuously aligns the policy with its evolving knowledge boundary. By eliminating the need for offline dataset relabeling and allowing the knowledge estimate to update throughout training, \rewardname{} not only reduces computational overhead but also fosters precise and generalizable abstention behavior.

\subsection{Two-Stage RL Training Strategy} 
\label{sec:two-stage}

Drawing on insights from Section \ref{sec:pilot}, we observe that: early in training, models are driven by the predominance of F\&U groups into an ``\textit{abstention trap}'', which prevents accuracy gains and can even degrade performance. 
To prevent this pitfall and ensure robust learning, we design a \textbf{Two-Stage RL Training Strategy}, which decouples the objectives of \textit{knowledge boundary} and \textit{abstention calibration} across two consecutive training stages.

\paragraph{Stage I: Knowledge Boundary Exploration.} 
This stage aims to maximize the elicitation and reinforcement of the model's factual knowledge, thereby establishing a high baseline of answer accuracy. 
To achieve this, we employ a \textbf{mixed-reward policy}: during training, the dataset is divided into two parts according to a ratio $\alpha$. The $\alpha$ proportion of data is trained using a simple binary reward ($r_{\text{pos}}=+1.0$, $r_{\text{neg}}=r_{\text{abs}}=0$), while the remaining data is trained using only the \rewardname{} mechanism described in Section~\ref{sec:method_reward}.
The binary reward acts as a safeguard: since both incorrect answers and abstentions receive zero reward, they yield zero advantage in the GRPO update, effectively nullifying gradients from groups that contain only errors or abstentions. 
This design circumvents the early-training dominance of such groups, which would otherwise push the policy toward excessive silence. Meanwhile, the \rewardname{}-trained portion ensures the model retains the ability to abstain. Consequently, this stage prioritizes confident exploration and knowledge recall, building a policy with a robust and strong factual grounding.

\paragraph{Stage II: Abstention Calibration.} 
Once the model's ability to answer questions within its knowledge boundary has been strengthened, the second stage focuses on precise calibration. 
In this stage, we transition to applying the \rewardname{} to \textit{all} training data solely. 
The goal is now to convert persistent errors---queries the model consistently answers incorrectly despite the Stage I exploration---into appropriate abstentions. Crucially, \rewardname{}'s conditional logic ensures this conversion is targeted: abstention is incentivized ($r_{\text{abs}}=+1.0$) only for queries where the model fails to generate any correct response within the sampled group ($\mathbb{I}_{\text{solvable}}=0$). 
For queries where the model demonstrates competence ($\mathbb{I}_{\text{solvable}}=1$), \rewardname{} continues to reward correct answers and penalize both errors and unnecessary abstentions. This selective pressure allows the model to refine its abstention boundary, suppressing hallucinations on genuinely unknown queries while largely preserving the high answer accuracy established in Stage I.

\section{Experiments}
\label{sec:experiments}
In this section, we present comprehensive experiments and detailed analyses to validate the effectiveness of our proposed \framename{}.

\subsection{Tasks and Evaluation Metrics}
\label{sec:datasets}

To thoroughly evaluate \framename{}, we perform experiments in multiple benchmarks, which are categorized into short-form knowledge tasks and reasoning tasks. This design aims to assess both factual precision and general problem-solving capabilities.

\noindent
\textbf{Short-form Knowledge Tasks.}
We primarily focus on knowledge-intensive tasks to evaluate the model's ability to provide accurate factual responses or abstain appropriately when uncertain.
We adopt NaturalQuestions (NQ)~\citep{NQ} as our in-distribution benchmark, using its training split for supervised fine-tuning and reinforcement learning, and the validation split for in-distribution evaluation.
To evaluate generalization, we include three out-of-distribution (OOD) datasets representing distinct distribution shifts.
First, we employ TriviaQA~\citep{Triviaqa} to assess performance on simpler general knowledge questions.
Second, we utilize BioASQ~\citep{BioASQ} to test whether the learned policy generalizes to a specialized biomedical domain.
Finally, we use ARC-Challenge~\citep{ARC}.
Unlike QA datasets, this dataset consists of single-choice questions, allowing us to evaluate abstention behavior under different question formats.

\noindent
\textbf{Reasoning Tasks.}
To ensure our method does not degrade general reasoning, we evaluate it on GSM8K~\citep{GSM8K} for multi-step mathematical reasoning and on BIG-Bench Hard (BBH)~\citep{BBH} for diverse and challenging reasoning tasks.

\noindent
\textbf{Evaluation Metrics.}
Following \citet{Rejection_Improves_Reliability}, 
we adopt a unified reliability score that balances accuracy and hallucination rate.
Let $T$, $U$, and $F$ denote the rates of correct answers, abstentions, and hallucinations, respectively (where $T+U+F=1$). 
The reliability metric is computed as: 
\begin{align*}
    Rely = (1-U) \times (1-F) + U \times T.
\end{align*}
This formulation can be interpreted as a dynamic trade-off: as the model answers more questions (higher $1-U$), the metric places greater weight on penalizing hallucinations ($F$); conversely, as it abstains more (higher $U$), it prioritizes the total yield of correct answers ($T$). This mechanism effectively penalizes both over-aggressive (high answer rate but high hallucination) and over-conservative (low answer rate and low correctness) strategies.

\begin{table*}[t]
\centering
\small
\setlength{\tabcolsep}{3.3pt}
\renewcommand{\arraystretch}{1.15}
\resizebox{\textwidth}{!}{
    \begin{tabular}{l cccc cccc cccc cccc}
    \toprule
     & \multicolumn{4}{c}{\textbf{NQ}} 
     & \multicolumn{4}{c}{\textbf{TriviaQA}} 
     & \multicolumn{4}{c}{\textbf{BioASQ}} 
     & \multicolumn{4}{c}{\textbf{ARC-C}} \\
    \cmidrule(lr){2-5}
    \cmidrule(lr){6-9}
    \cmidrule(lr){10-13}
    \cmidrule(lr){14-17}
    
    \quad Method 
    & \textbf{T} ($\uparrow$) & \textbf{U} ($-$) & \textbf{F} ($\downarrow$)  & \color{red!60!black}{\textbf{Rely} ($\uparrow$)}
    & \textbf{T} ($\uparrow$) & \textbf{U} ($-$) & \textbf{F} ($\downarrow$)  & \color{red!60!black}{\textbf{Rely} ($\uparrow$)}
    & \textbf{T} ($\uparrow$) & \textbf{U} ($-$) & \textbf{F} ($\downarrow$) & \color{red!60!black}{\textbf{Rely} ($\uparrow$)}
    & \textbf{T} ($\uparrow$) & \textbf{U} ($-$) & \textbf{F} ($\downarrow$)  & \color{red!60!black}{\textbf{Rely} ($\uparrow$)}\\
    \midrule
    \rowcolor{myyellow}
    \multicolumn{17}{c}{\textbf{Llama3.1-8B-Inst}} \\
    
    \quad Prompting 
    & 38.7 & 6.0 & 55.3 & 44.3
    & 71.6 & 1.3 & 27.1 & 72.9
    & 55.8 & 0.3 & 43.9 & 56.1
    & 81.7 & 0.3 & 18.0 & 82.0\\
    
    \quad SFT 
    & 40.0 & 0.0 & 60.0 & 40.0
    & 70.2 & 0.0 & 29.8 & 70.2
    & 53.4 & 0.0 & 46.6 & 53.4
    & 77.6 & 0.0 & 22.4 & 77.6\\
    
    \quad R-Tuning 
    & 24.8 & 56.8 & 18.4 & 49.3
    & 55.4 & 35.9 & 8.7 & 78.4
    & 27.9 & 65.6 & 6.5 & 50.5
    & 10.3 & 87.8 & 1.9 & 21.0\\
    
    \quad RLKF 
    & 28.7 & 30.7 & 40.6 & 50.0
    & 57.6 & 23.2 & 19.2 & 75.4
    & 43.3 & 27.6 & 29.1 & 63.3
    & 54.9 & 5.2 & 39.9 & 59.8\\
    \hdashline[2pt/3pt]
    \quad Binary
    & 56.3 & 0.0 & 43.7 & 56.3
    & 76.8 & 0.0 & 23.2 & 76.8
    & 66.0 & 0.0 & 34.0 & 66.0
    & 82.4 & 0.0 & 17.6 & 82.4\\
    
    \quad Ternary {\scriptsize (TruthRL)}
    & 34.9 & 53.6 & 11.5 & 59.8
    & 61.4 & 32.3 & 6.3& 83.3
    & 39.0 & 57.0 & 4.0& 63.5
    & 71.9 & 18.1 & 10.0& 86.7\\\rowcolor{blue!5} \quad \textbf{Ours} 
    & 42.7 & 40.6 & 16.7 & \textbf{66.8}
    & 67.9 & 23.5 & 8.6 & \textbf{85.9}
    & 57.7 & 32.8 & 9.5 & \textbf{79.7}
    & 80.0 & 9.1 & 10.9 & \textbf{88.3}\\\rowcolor{blue!5}$\Delta$ Compared to Prompting
    & +4.0 & +34.6 & -38.6 & \color{green!60!black}{+22.5}
    & -3.7 & +22.2 & -18.5 & \color{green!60!black}{+13.0}
    & +1.9 & +32.5 & -34.4 & \color{green!60!black}{+23.6}
    & -1.7 & +8.8 & -7.1 & \color{green!60!black}{+6.3}\\
    
    \midrule
    \rowcolor{myyellow}
    \multicolumn{17}{c}{\textbf{Qwen2.5-7B-Inst}} \\
    
    \quad Prompting 
    & 23.0 & 32.5 & 44.5 & 44.9
    & 49.7 & 17.0 & 33.3 & 63.8
    & 43.2 & 21.8 & 35.0 & 60.2
    & 88.0 & 0.9 & 11.1 & 88.9\\
    
    \quad SFT 
    & 30.1 & 0.0 & 69.9 & 30.1
    & 60.3 & 0.0 & 39.7 & 60.3
    & 49.4 & 0.0 & 50.6 & 49.4
    & 87.0 & 0.0 & 13.0 & 87.0\\
    
    \quad R-Tuning 
    & 9.3 & 87.4 & 3.3 & 20.3
    & 35.4 & 60.1 & 4.5 & 59.4
    & 22.7 & 75.2 & 2.1 & 41.3
    & 52.6 & 45.0 & 2.4 & 77.4\\

    \quad RLKF 
    & 24.9 & 43.2 & 31.9 & 49.4
    & 41.7 & 1.7 & 56.6 & 43.4
    & 35.3 & 7.0 & 57.7 & 41.8
    & 85.1 & 4.2 & 10.7 & 89.1\\
    \hdashline[2pt/3pt]
    \quad Binary
    & 37.0 & 0.0 & 63.0 & 37.0
    & 64.2 & 0.0 & 35.8 & 64.2
    & 60.7 & 0.0 & 39.3 & 60.7
    & 87.6 & 0.0 & 12.4 & 87.6\\
    
    \quad Ternary {\scriptsize (TruthRL)}
    & 11.9 & 83.0 & 5.1 & 26.0
    & 39.2 & 52.2 & 8.6 & 64.2
    & 25.2 & 67.8 & 7.0 & 47.0
    & 83.6 & 8.2 & 8.2 & \textbf{91.1}\\\rowcolor{blue!5} \quad \textbf{Ours} 
    & 26.3 & 49.5 & 24.2 & \textbf{51.3}
    & 51.7 & 30.9 & 17.4 & \textbf{73.1}
    & 43.0 & 35.2 & 21.8 & \textbf{65.8}
    & 87.2 & 2.3 & 10.5 & 89.4 \\\rowcolor{blue!5}$\Delta$ Compared to Prompting
    & +3.3 & +17.0 & -20.3 & \color{green!60!black}{+6.4}
    & +2.0 & +13.9 & -15.9 & \color{green!60!black}{+9.3}
    & -0.2 & +13.4 & -13.2 & \color{green!60!black}{+5.6}
    & -0.8 & +1.4 & -0.6 & \color{green!60!black}{+0.5}\\
    
    \bottomrule
    \end{tabular}
}
\vspace{2mm}
\caption{
Experimental results (\%) on in-distribution and three out-of-distribution datasets. \textbf{T}, \textbf{U}, and \textbf{F} denote the rates of \textbf{correct answers}, \textbf{abstentions}, and \textbf{hallucinations}, respectively. \textcolor{red!60!black}{\textbf{Rely}} is a unified reliability score that balances accuracy and hallucination rate.
The $\Delta$ row reports the performance difference between our method and the Prompting baseline. \textcolor{green!60!black}{Green} numbers indicates improvement. Best performance is highlighted in \textbf{bold}.
}
\label{tab:main_results}
\end{table*}

\subsection{Baselines}
\label{sec:baselines}
We compare \framename{} against several representative baselines from three categories. Specifically,
\textit{Inference-Only}: \textbf{Prompting} directly prompts the base model to answer or abstain when uncertain, without any further training.
\textit{Supervised Fine-Tuning}:
(1) \textbf{SFT} performs standard supervised fine-tuning on the original dataset, treating the ground-truth answers as the training targets.
(2) \textbf{R-Tuning}~\cite{R-Tuning} is an abstention-aware method that relabels uncertain examples as ``I don't know'' based on offline consistency probing, then fine-tunes on the relabeled dataset.
\textit{Reinforcement Learning}:
(1) \textbf{RLKF}~\cite{Rejection_Improves_Reliability} aligns the model using DPO with preferences derived from offline sampling consistency to align answer, abstention, and error preferences.
(2) \textbf{Binary} applies GRPO with a naive binary reward, assigning a reward of $+1$ for correct answers and $0$ for any other response.
(3) \textbf{Ternary} (TruthRL)~\citep{truthrl} utilizes GRPO with a static ternary reward ($+1$ for correct, $0$ for abstention, $-1$ for incorrect) to explicitly encourage abstention.

\subsection{Implementation Details}
We evaluate our method on two widely-used open-source large language models: Llama-3.1-8B-Instruct~\citep{Llama3} and Qwen2.5-7B-Instruct~\citep{Qwen2.5}. For our method, the mixing ratio $\alpha$ for the binary reward in Stage~I is set to 50\%. 
All models are trained on a randomly sampled subset of 19,200 examples from the NQ training split. For each generated response during the training and evaluation phases, we employ Llama-3.3-70B-Instruct as the judge model to evaluate its correctness against the gold answer via zero-shot prompting. Additional implementation details are provided in Appendix~\ref{appendix:implementation}.

\subsection{Main Results}

\noindent
\textbf{\framename{} Achieves a Superior Accuracy-Abstention Balance.}
Table~\ref{tab:main_results} presents the performance comparisons on the in-distribution NQ dataset and three OOD datasets.
Methods that ignore abstention, such as SFT and Binary Reward, achieve high accuracy but suppress abstention completely, leading to excessive hallucinations.
Existing abstention-aware approaches like R-Tuning, RLKF, and Ternary Reward effectively increase abstention rates but significantly compromise accuracy compared to the base model (Prompting).
In contrast, \framename{} successfully overcomes this limitation and achieves the highest $Rely$ scores across all evaluated datasets and models.
Specifically, on NQ, our method consistently improves performance over the Prompting baseline: it reduces the hallucination rate by \textbf{38.6\%} for Llama3.1 and \textbf{20.3\%} for Qwen2.5, while simultaneously improving accuracy by \textbf{4.0\%} and \textbf{3.3\%}, respectively.
This indicates that our method effectively teaches the model to abstain from unknown queries without suppressing its ability to answer solvable ones.

\noindent
\textbf{\framename{} Generalizes Robustly to OOD Datasets.}
We further evaluate the transferability of the learned abstention policy on diverse unseen datasets, including TriviaQA, BioASQ, and ARC-C.
As shown in Table~\ref{tab:main_results}, methods relying on offline data construction (SFT, R-tuning, RLKF) exhibit limited generalization.
Across diverse domains and formats, these approaches either suffer from severe accuracy degradation or fail to reduce hallucinations compared to the prompting baseline.
Regarding RLVR-based baselines, the Binary baseline consistently lacks abstention capability across all datasets.
Meanwhile, the Ternary baseline suffers from significant accuracy drops on OOD tasks across both model families, such as a 16.8\% and 18.0\% decrease on BioASQ for Llama-3.1 and Qwen2.5, respectively.
In contrast, \framename{} demonstrates robust generalization.
It consistently maintains high accuracy with minimal performance drops and even achieves accuracy gains, such as \textbf{1.9\%} on BioASQ for Llama3.1 and \textbf{2.0\%} on TriviaQA for Qwen2.5.
This indicates that our method learns a generalized capability to abstain appropriately rather than overfitting to training patterns.

\begin{table}[t]
\scriptsize
\centering
\renewcommand{\arraystretch}{1.15}
\setlength{\tabcolsep}{4pt}

\begin{tabular}{lcccccccccccc}
\toprule
\rowcolor{myyellow}
& \multicolumn{6}{c}{\textbf{Llama3.1-8B-Instruct}} 
& \multicolumn{6}{c}{\textbf{Qwen2.5-7B-Instruct}} \\
\cmidrule(lr){2-7} \cmidrule(lr){8-13}

\multirow{2}{*}{\textbf{Method}}
& \multicolumn{3}{c}{GSM8K} & \multicolumn{3}{c}{BBH}
& \multicolumn{3}{c}{GSM8K} & \multicolumn{3}{c}{BBH} \\
\cmidrule(lr){2-4} \cmidrule(lr){5-7}
\cmidrule(lr){8-10} \cmidrule(lr){11-13}

& \textbf{T} ($\uparrow$) & \textbf{U} ($-$) & \textbf{F} ($\downarrow$) & \textbf{T} ($\uparrow$) & \textbf{U} ($-$) & \textbf{F} ($\downarrow$) & \textbf{T} ($\uparrow$) & \textbf{U} ($-$) & \textbf{F} ($\downarrow$) & \textbf{T} ($\uparrow$) & \textbf{U} ($-$) & \textbf{F} ($\downarrow$) \\
\midrule

Prompting 
& 83.7 & 0.5 & 15.8 & 64.3 & 2.1 & 33.6
& 89.1 & 0.5 & 10.4 & 67.5 & 5.8 & 26.7 \\

Binary
& 85.2 & 0.0 & 14.8 & 63.2 & 0.0 & 36.8
& 91.4 & 0.0 & 8.6 & 71.5 & 0.1 & 28.4 \\

Ternary
& 85.5 & 3.0 & 11.5 & 53.4 & 28.1 & 18.5
& 88.5 & 3.0 & 8.5 & 61.5 & 15.6 & 22.9 \\

\rowcolor{blue!5}
\textbf{Ours}
& 86.1 & 1.3 & 12.6 & 62.5 & 11.3 & 26.2
& 90.4 & 0.4 & 9.2 & 67.8 & 7.4 & 24.8 \\

\bottomrule
\end{tabular}

\caption{Performance (\%) comparison on GSM8K and BBH using Llama-3.1-8B-Instruct and Qwen2.5-7B-Instruct.}
\label{tb:reasoning_results}
\end{table}

\noindent
\textbf{\framename{} Preserves Reasoning Capabilities.}
We further examine whether training on knowledge tasks compromises reasoning abilities using GSM8K and BBH.
Table~\ref{tb:reasoning_results} shows that \framename{} consistently maintains or improves accuracy compared to the prompting baseline while effectively reducing hallucinations.
Regarding baselines, the Binary baseline completely fails to abstain in reasoning contexts.
The Ternary baseline shows instability, notably suffering a significant accuracy drop of 10.9\% on BBH for Llama3.1 and 6.0\% for Qwen2.5.
Overall, these results demonstrate that \framename{} achieves robust generalization without degrading the model's core reasoning intelligence.


\noindent
\textbf{Avoiding the ``\textit{Abstention Trap}''.}
\label{sec:trap}
As discussed in our pilot study, the dominance of F\&U groups often drives models into an ``\textit{abstention trap}'' early in training.
To validate the effectiveness of our method, we analyze the training dynamics of the binary baseline, the ternary baseline and our method across different batch sizes (64, 128, and 256).
Figure~\ref{fig:early_training_dynamics} illustrates the evolution of correct and abstention rates during the early steps.
The Ternary baseline consistently experiences a significant drop in accuracy early in the process.
This instability is most severe at a batch size of 64, where the model rapidly converges to 100\% abstention.
At this point, the model loses all gradient signals and ceases to learn.
In contrast, our method bypasses this failure mode.
Across all batch sizes, our approach achieves accuracy growth comparable to the Binary baseline while retaining a significantly higher abstention rate.
This indicates that the model effectively explores its knowledge boundary without sacrificing abstention awareness.
This stability avoids the early collapse seen in the ternary baseline and is crucial for achieving high final accuracy and robust performance.

\subsection{Analysis}
\begin{figure}[t]
  \centering
  \begin{minipage}{0.48\textwidth}
    \includegraphics[width=\linewidth]{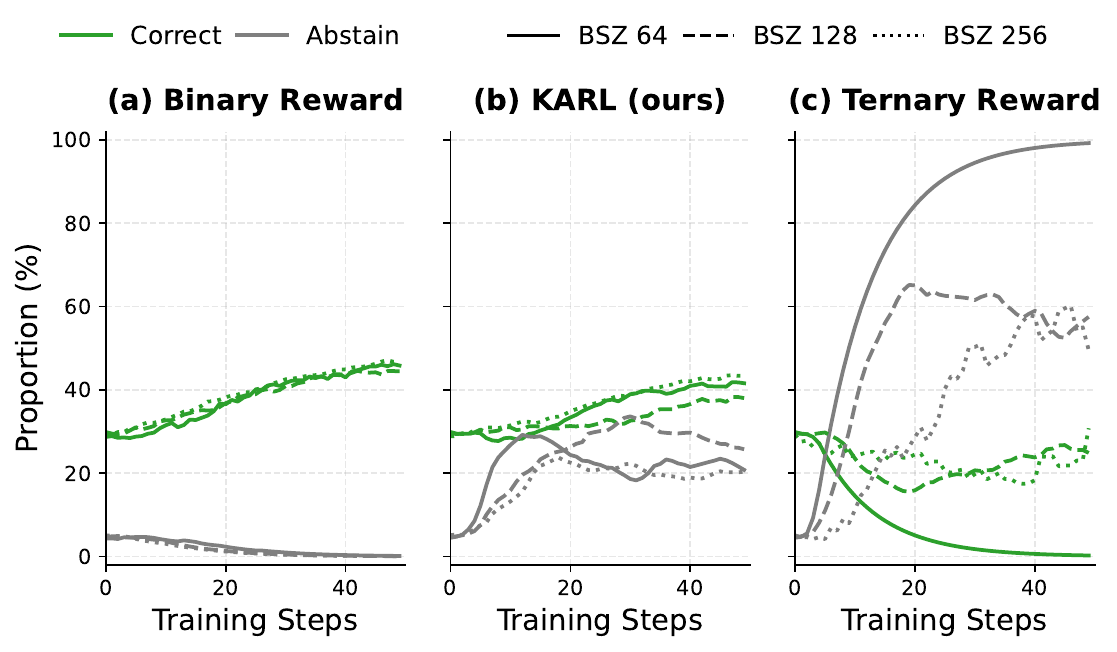}
    \caption{Early training dynamics of \textbf{Correct} (green) and \textbf{Abstain} (gray) response proportions on the NQ dataset.
        The subplots compare (a) Binary Reward, (b) \framename{} (Ours), and (c) Ternary Reward across three batch sizes (64, 128, 256).}
    \label{fig:early_training_dynamics}
  \end{minipage}
  \hfill
  \begin{minipage}{0.48\textwidth}
    \small
    \centering
    \renewcommand{\arraystretch}{1.15}
    \resizebox{1\linewidth}{!}{
    \begin{tabular}{lcccc}
    \toprule
    \textbf{Setting} & \textbf{T ($\uparrow$)} & \textbf{U ($-$)} & \textbf{F ($\downarrow$)} & \color{red!60!black}{\textbf{Rely} ($\uparrow$)}\\
    \midrule
    Prompting & 38.7 & 6.0 & 55.3 & 44.3\\
    \midrule
    \rowcolor{myyellow}
    \multicolumn{5}{l}{\textit{\textbf{Ablation on Strategy}}} \\
    \quad w/o Stage II & 50.6 & 18.0 & 31.4 & 65.4 \\
    \midrule
    \rowcolor{myyellow}
    \multicolumn{5}{l}{\textit{\textbf{Ablation on Mixing Ratio $\alpha$ in Stage I}}} \\
    \quad $\alpha = 0\%$ (Pure \rewardname{})  & 36.0 & 50.0 & 14.0 & 61.0\\
    \quad $\alpha = 25\%$  & 42.4 & 35.5 & 22.1 & 65.3\\
    \quad $\alpha = 50\%$  & 42.7 & 40.6 & 16.7 & \textbf{66.8}\\
    \quad $\alpha = 75\%$  & 39.3 & 47.0 & 13.7 & 64.2\\
    \quad $\alpha = 100\%$ (Pure Binary) & 55.5 & 0.0 & 44.5 & 55.5\\
    \bottomrule
    \end{tabular}
    }
    \captionof{table}{Ablation study on NQ using Llama-3.1-8B-Instruct.
    We evaluate the impact of the two-stage strategy by removing the second stage (w/o Stage II) and analyze the impact of mixing ratio ($\alpha$ ) of binary reward data in Stage I. Best performance is highlighted in \textbf{bold}.}
    \label{tab:ablation_all}
      \end{minipage}
\end{figure}

\noindent
\textbf{Ablation Study.}
We conduct an ablation study in Table~\ref{tab:ablation_all} to verify the contribution of each component in our framework. Detailed ablation results across all datasets are provided in Appendix~\ref{appendix: ablation_all_datasets}.



\textit{Necessity of Two-Stage Training.} We assess the necessity of the two-stage strategy.
Removing the second stage results in the highest accuracy but a significantly lower abstention rate compared to the full method.
This happens because the primary goal of Stage II is to use the full \rewardname{} mechanism to convert the remaining incorrect answers into abstentions.
Without this calibration stage, the model remains in an exploratory state, leading to higher hallucination rates and compromised  reliability.

\textit{Impact of $\alpha$ in Stage~I.} We investigate the impact of the binary reward mixing ratio $\alpha$ in Stage I.
When $\alpha$ is set to 0\%, the method degrades to single-stage training with pure \rewardname{}.
This leads to the lowest accuracy, as the penalties in \rewardname{} suppress the exploration of knowledge boundary during the early stages.
Conversely, setting $\alpha$ to 100\% means the model is trained exclusively with binary rewards in Stage I.
Under this setting, the model completely loses its ability to abstain in Stage I and fails to recover in Stage II because it cannot generate any abstention responses during rollout, resulting in a zero abstention rate.
We find that a mixing ratio of 50\% achieves the optimal balance and the highest $Rely$ score, effectively eliciting knowledge while preserving abstention capability for the final calibration stage.


\textbf{Discussion.}
To further evaluate the stability and practical requirements of \framename{} under diverse training configurations, we investigate several key factors that might influence its alignment performance. Detailed analyses are provided in Appendix \ref{appendix:discussion}, including: (1) an investigation into the sensitivity of the GRPO rollout size ($G$) in Appendix \ref{app: ablation on Group size}; (2) an analysis of the model's performance when trained on datasets of varying difficulty levels in Appendix \ref{app: ablation on Data Difficulty}; and (3) a study on the optimal timing for the transition between Stage I and Stage II in Appendix \ref{app: ablation on Transition timing}.
We also present a case study in Appendix \ref{sec: case study} to further demonstrate the advantages of our method.

Despite the effectiveness of KARL, several limitations remain.
First, our current reward mechanism is still coarse, as it only distinguishes among three discrete states: correct, incorrect, and abstention.
As a result, sampled groups that yield uniform rewards produce zero advantage and contribute no gradient updates.
Furthermore, this discrete structure is less suited for long-form or open-ended generation settings, which inherently require more granular scoring mechanisms to capture nuances beyond simple correctness. Future work could investigate finer-grained reward designs to provide denser learning signals and broader applicability.
Second, our two-stage curriculum currently relies on a fixed schedule to transition between exploration and calibration.
A more advanced approach would be to dynamically toggle between these phases based on the real-time performance of the model during training, which we leave as a direction for future research.
Finally, similar to most RL methods, the effectiveness of our framework on challenging tasks is limited when the training data is too easy. Future research should consider incorporating training data with sufficient difficulty to allow the model to explore its knowledge boundary and develop more robust learning mechanisms that can effectively navigate the sparse feedback signals.

\section{Conclusion}
\label{sec:bibtex}
In this paper, we propose \framename{}, a novel framework that continuously aligns an LLM's abstention behavior with its evolving knowledge boundary. \framename{} introduces two core innovations: a \textit{Knowledge-Boundary-Aware Reward} that performs online knowledge estimation using within-group response statistics, dynamically rewarding correct answers or guided abstention; and a \textit{Two-Stage RL Training Strategy} that first explores knowledge boundary and bypasses the ``\textit{abstention trap}'', and subsequently converts incorrect answers beyond the knowledge boundary into abstentions without sacrificing accuracy. Experimental results demonstrate that \framename{} achieves a superior trade-off between accuracy and abstention, significantly suppressing hallucinations while maintaining high answer accuracy across diverse in-distribution and out-of-distribution scenarios.

\bibliography{custom}
\bibliographystyle{colm2026_conference}

\appendix  

\section{Related Works}
\subsection{Supervised Fine-Tuning for Abstention.}
Recent advancements suggest that post-training remains an effective stage for mitigating persistent hallucinations. \citep{h-neurons, KLCF}.

Supervised approaches teach LLMs to abstain by fine-tuning on datasets containing explicit abstention labels. 
A representative method, R-Tuning~\citep{R-Tuning}, estimates the model's knowledge boundary via multi-sample consistency probing and relabels responses to uncertain questions with ``I don't know''. 

Subsequent Refusal-Aware Instruction Tuning (RAIT) strategies extend this paradigm by constructing refusal data through diverse mechanisms, such as filtering low-accuracy samples~\citep{AlignmentforHonesty, CanAIKnowWhatTheyDontKnow,GATEAU}, incorporating unanswerable QA pairs~\citep{Laboratory-ScaleAI, wen2024Knowyourlimits}, or synthesizing abstention scenarios via prompting~\citep{Brahman2024SayingNo,DontJustSayIdontknow,Aligning, luo-etal-2025-lets}.
These methods train models to map out-of-knowledge inputs directly to standardized abstention responses.
While straightforward, SFT-based approaches fundamentally rely on memorizing specific refusal patterns rather than learning dynamic decision boundaries. 
Consequently, they often lead to over-refusals, causing models to refuse benign questions and generalize poorly to out-of-distribution domains~\citep{Understanding_RL}.
Furthermore, constructing such datasets incurs high computational costs. Since knowledge boundaries are intrinsic and vary across models, abstention labels must be customized via extensive probing for each specific model, preventing efficient transferability.
Although recent efforts like GRAIT~\citep{GRAIT} attempt to mitigate over-refusal via gradient-based sample weighting, SFT methods remain limited by the static and model-specific nature of their supervision signals.

\subsection{Reinforcement Learning for Abstention.}
Existing RL-based approaches for abstention can be broadly grouped by the type of supervision signal they rely on.
One line of work adopts \emph{preference-based} RL, where models are aligned using PPO or DPO against a reward model or a static set of preference pairs in which abstention is preferred over hallucination~\citep{HH-RLHF,LLMsKnowWhatTheyKnow,Rejection_Improves_Reliability}.
Despite their effectiveness, these methods face scalability challenges similar to SFT, as they require large amounts of carefully curated preference data tailored to abstention scenarios.

To reduce reliance on preference annotations, recent work has shifted toward \emph{outcome-based} RL, commonly referred to as Reinforcement Learning with Verifiable Rewards (RLVR).
These methods leverage automatic verification signals to directly optimize model behavior without training a separate reward model~\citep{DeepSeekMath,Deepseek-r1}.
While computationally efficient, standard RLVR objectives tend to amplify model confidence, reducing the likelihood of abstention and exacerbating hallucinations~\citep{NIPS-RL,TheHallucinationTax}.
To mitigate this, recent works have attempted to refine the reward design within the RLVR framework~\citep{Teaching}.
For instance, ConfClip~\citep{ConfClip} integrates confidence scores into the GRPO reward to encourage high-confidence correctness and penalize high-confidence hallucination, while KLCF~\citep{KLCF} penalizes responses that exceed the model's knowledge boundary.
Nevertheless, these methods lack an explicit mechanism for abstention actions, often resulting in persistent hallucination rates.
Addressing this, TruthRL~\citep{truthrl} incorporates a ternary reward scheme within GRPO to explicitly distinguish between correct answers, hallucinations, and abstentions.
Similarly, MASH~\citep{MASH} trains models to seek external knowledge when they are not sure, then converts search actions into abstentions during inference.
While these explicit methods effectively reduce hallucinations, they still suffer from over-refusals and degrade the model's overall accuracy and utility.
Thus, how to eliminate complex data construction while maintaining a balance between model utility and hallucination reduction still remains underexplored.

\section{Reinforcement Learning Protocol}
\label{app:GRPO}

For the RL training of LLMs, policy optimization methods such as PPO~\citep{ppo} and GRPO~\citep{DeepSeekMath} have been widely studied. 
GRPO is especially attractive in verifiable-reward settings as it avoids a separate value network or reward model by estimating advantages directly from reward statistics of sampled responses. 
Therefore, we adopt GRPO as our backbone RL framework.

Specifically, for each input query $q$, GRPO samples a group of $G$ responses $\{o_1, \dots, o_G\}$ from the policy before the update, i.e., $\pi_{\theta_{\text{old}}}$. It then optimizes the policy by maximizing a surrogate objective that encourages actions with above-average rewards while penalizing deviation from a reference policy $\pi_{\text{ref}}$ via a KL-divergence term. 
The GRPO objective is:
\begin{equation}		
\label{grpo_1}
		\begin{aligned}
            \mathcal{J}_\text{GRPO}(\theta) &= \mathbb{E}_{q, \{o_i\} \sim \pi_{\theta_{old}}} \left[ \frac{1}{G} \sum_{i=1}^G \mathcal{L}_i - \beta \mathbb{D}_{KL}(\pi_{\theta} || \pi_{ref}) \right],
		\end{aligned}
\end{equation}
\begin{equation}		
\label{grpo_2}
		\begin{aligned}
            \mathcal{L}_i &= \min \left( w_i \hat{A}_i, \text{clip}(w_i, 1 - \epsilon, 1 + \epsilon) \hat{A}_i \right),
		\end{aligned}
\end{equation}
where $w_i = \frac{\pi_{\theta}(o_i|q, c)}{\pi_{\theta_{\text{old}}}(o_i|q, c)}$, $\epsilon$ and $\beta$ are hyperparameters that constrain the policy update and control the divergence penalty, respectively.
Crucially, $A_i$ represents the advantage value for the $i$-th output, which is derived from the group-wise statistics:
\begin{equation}
\label{eq:advantage}
\hat{A}_i = \frac{r_i - \mu(\{r_1, \cdots, r_G\})}{\sigma(\{r_1, \cdots, r_G\}) + \delta},
\end{equation}
where $r_i$ is the reward for output $o_i$, $\mu$ and $\sigma$ denote the mean and standard deviation over the group, and $\delta$ is a small constant.

\section{Implementation Details}
\label{appendix:implementation}

\subsection{Pilot Study Implementation Details}
\label{app:pilot_details}

\noindent\textbf{Training Dynamics Setup.}
The experimental setup for the pilot study follows the general training protocols described in Appendix~\ref{app:Training_Details}, with the following specific configurations:

We utilize Llama-3.1-8B-Instruct as the base model, fine-tuned on the NQ Open dataset. Unlike the main experiments where we consider a range of batch sizes, we fix the global batch size at \textbf{128} for all pilot experiments to ensure a controlled comparison. All other hyperparameters remain identical to the main configuration.
Regarding reward structures, the Binary setup assigns a reward of $1$ for correct answers and $0$ for any other response, whereas the Ternary setup assigns 1 for correctness, 0 for abstention, and -1 for errors.

\noindent\textbf{Rollout Distribution Analysis.}
To analyze the mechanics behind the abstention trap, we conducted rollout experiments using Llama-3.1-8B-Instruct and Qwen2.5-7B-Instruct on the NQ and TriviaQA datasets.
For each query, we generate 8 responses using probabilistic decoding parameters (\texttt{temperature=1.0}, \texttt{top\_k=50}, \texttt{top\_p=0.9}). 
We classified each response as Correct, Incorrect, or Abstention and categorized the resulting groups based on their composition.
In our analysis of the gradient contribution, we filtered out groups that do not influence the abstention policy.
Specifically, we excluded groups consisting of a single response type as they yield zero advantage.
We also excluded groups containing only correct and incorrect answers as they do not directly drive the learning of abstention.
The distribution shown in Figure~\ref{fig:rollout_distribution} reflects the remaining groups that actively influence abstention.

\subsection{Training Details}
\label{app:Training_Details}
All models are trained on 8 NVIDIA A800 GPUs with 40GB memory using full-parameter fine-tuning and trained for 1 epoch.

For GRPO, we implement our training pipeline using \texttt{verl} framework~\citep{verl}. To manage memory usage during full-parameter fine-tuning, we utilize DeepSpeed~\citep{deepspeed} with ZeRO-3 offload and enable gradient checkpointing.
We employ a constant learning rate of $1\text{e-}6$, a KL divergence coefficient $\beta$ of 0.001, and perform a hyperparameter search for batch sizes over \{64, 128, 256\}.

We use vLLM~\citep{vllm} engine for LLM rollouts in GRPO with a tensor parallel size of 2 and GPU memory utilization of 0.8. For each query, we sample $G=8$ responses. The sampling parameters are set to a temperature of 1.0 and top-p of 1.0. For short-form knowledge tasks, we train without Chain-of-Thought (CoT) and restrict the maximum response length to 64 tokens, using the inference prompt shown in Figure~\ref{fig:prompt_without_CoT_qa}. For reasoning tasks, we encourage CoT generation and set the maximum response length to 512 tokens, using the inference prompt shown in Figure~\ref{fig:prompt_with_CoT_qa}. The prompt template used by the judge model (Llama3.3-70B-Instruct) for GRPO training rewards is shown in Figure~\ref{fig:judge_prompt}.

For SFT training and R-Tuning, we use a learning rate of $5 \times 10^{-6}$ and a batch size of 8. For DPO training, we use a learning rate of $3\times 10^{-6}$ and a batch size of 8. All these training runs adopted a cosine learning rate scheduler with $3\%$ warmup steps and are conducted for 1 epoch over the dataset.


\subsection{Evaluation Details}
We use vLLM and apply greedy decoding during evaluation. All results are evaluated in a 0-shot setting.
To ensure precise metric calculation, we first use rule-based string matching to detect abstention expressions (e.g., ``I don't know''). Responses that include these expressions are classified as Abstention.
Only non-refusal responses are passed to the judge model (Llama-3.3-70B-Instruct) to determine factual correctness using the prompt in Figure~\ref{fig:judge_prompt}.

This separation is crucial for preventing reward hacking.
We observe that relying solely on the LLM to classify all three states (Correct, Incorrect, Abstention) encourages the trained model to generate ambiguous responses.
Such ambiguity allows the model to be interpreted as correct when the answer is included, or as uncertain otherwise, thereby bypassing error penalties.
Furthermore, we employ the LLM judge for correctness instead of a rule-based judge because reference answers in open-domain QA datasets often have limited coverage of valid answer expressions, a limitation also noted in prior work~\citep{truthrl}.

\section{Discussion}
\label{appendix:discussion}

\subsection{Case Study}
\label{sec: case study}
As shown in the examples in Table \ref{tab:case}, baseline methods frequently fail due to hallucinations or unnecessary abstention. 
In contrast, our approach effectively utilizes the reasoning capabilities enhanced by RL.
This enables the model to answer challenging questions correctly without being overly cautious, achieving a more robust balance between factual accuracy and abstention.

\begin{table}[t]
    \scriptsize
    \centering
    \resizebox{\linewidth}{!}{
        \begin{tabular}{m{\hsize}}
        
        \specialrule{1pt}{1pt}{2pt}
        \tightbox{tabcolor4}{\textbf{NQ}} \\ 
        Question: What type of database is library literature and information science? \\
        
        \specialrule{0pt}{0pt}{1pt}
        \cdashline{1-1}[0.8pt/2pt]
        \specialrule{0pt}{0pt}{2pt}
        
        \tightbox{tabcolor6}{\textbf{Prompting}}: Metadata database \textit{\textbf{(\textcolor{red!60!black}{F})}} \\

        \specialrule{0pt}{0pt}{1pt}
        \cdashline{1-1}[0.8pt/2pt]
        \specialrule{0pt}{0pt}{2pt}
        \tightbox{tabcolor6}{\textbf{Ternary}}:
        Library Literature and Information Science is a database that indexes and abstracts articles and other publications in the field of library and information science. I don't know its specific database type.
        \textbf{\textit{(U)}} \\        
        \specialrule{0pt}{0pt}{1pt}
        \cdashline{1-1}[0.8pt/2pt]
        \specialrule{0pt}{0pt}{2pt}
        \tightbox{tabcolor1}{\textbf{Ours}}:
         Library Literature and Information Science is a database that indexes articles and publications related to library science, often a bibliographic or index database.
        \textbf{\textit{(\textcolor{green!60!black}{T})}} \\   
        \specialrule{1pt}{0pt}{1pt}
        \tightbox{tabcolor4}{\textbf{ARC-C}} \\ 
        Question: Farmers in Wyoming were concerned because some of their chickens were being preyed upon by hawks that lived in areas around their ranches. The farmers grouped together and hunted the hawks until they were no longer in their area. Which would most likely happen next? Options: A. The chicken population would go down. B. Populations of mice and rats would increase. C. Another bird of prey would replace the hawk. D. The chickens would have a lower rate of disease. \\
        
        \specialrule{0pt}{0pt}{1pt}
        \cdashline{1-1}[0.8pt/2pt]
        \specialrule{0pt}{0pt}{2pt}
        
        \tightbox{tabcolor6}{\textbf{Prompting}}: C. Another bird of prey would replace the hawk. \textit{\textbf{(\textcolor{red!60!black}{F})}} \\

        \specialrule{0pt}{0pt}{1pt}
        \cdashline{1-1}[0.8pt/2pt]
        \specialrule{0pt}{0pt}{2pt}
        \tightbox{tabcolor6}{\textbf{Ternary}}:
        The chicken population would likely increase due to the removal of a predator, but I don't know the exact outcome.
        \textbf{\textit{(U)}} \\        
        \specialrule{0pt}{0pt}{1pt}
        \cdashline{1-1}[0.8pt/2pt]
        \specialrule{0pt}{0pt}{2pt}
        \tightbox{tabcolor1}{\textbf{Ours}}:
         The removal of hawks would likely lead to an increase in populations of mice and rats, as their natural predator is removed, so B is a likely outcome.
        \textbf{\textit{(\textcolor{green!60!black}{T})}} \\   
        \specialrule{1pt}{0.5pt}{0pt}   
        \end{tabular}}
\caption{Case examples from the NQ and ARC-C datasets comparing responses generated by the Prompting baseline, Ternary baseline, and our method.
\textbf{(\textit{\textcolor{green!60!black}{T}})}, \textbf{(\textit{\textcolor{red!60!black}{F}})}, and \textbf{(\textit{U})} denote correct answers, incorrect answers, and abstentions, respectively. Models are implemented on Llama-3.1-8B-Instruct.}
    
\label{tab:case}
\end{table}

\subsection{Ablation on Group Size $G$}
\label{app: ablation on Group size}

\begin{figure}[t]
  \centering
  \begin{minipage}{0.43\textwidth}
    \small
    \centering
    \renewcommand{\arraystretch}{1.15}
    \resizebox{1\linewidth}{!}{
    \begin{tabular}{lcccc}
    \toprule
    \textbf{Setting} & \textbf{T ($\uparrow$)} & \textbf{U ($-$)} & \textbf{F ($\downarrow$)} & \color{red!60!black}{\textbf{Rely} ($\uparrow$)}\\
    \midrule
    \rowcolor{myyellow}
    \multicolumn{5}{l}{\textit{\textbf{Ablation on Group Size $G$}}} \\
    $G = 4$ & 32.2 & 56.8 & 12.0 & 56.3 \\
    $G = 8$ & 42.7 & 40.6 & 16.7 & \textbf{66.8} \\
    $G = 16$ & 41.8 & 42.2 & 16.0 & 66.2 \\
    \bottomrule
    \end{tabular}}
    \captionof{table}{Ablation study on the group size $G$ during the GRPO rollout process.}
    \label{tab:ablation_g}
      \end{minipage}
  \hfill
  \begin{minipage}{0.53\textwidth}
    \small
    \centering
    \renewcommand{\arraystretch}{1.15}
    \resizebox{1\linewidth}{!}{
    \begin{tabular}{lcccc}
    \toprule
    \textbf{Training Data} & \textbf{T ($\uparrow$)} & \textbf{U ($-$)} & \textbf{F ($\downarrow$)} & \color{red!60!black}{\textbf{Rely} ($\uparrow$)}\\
    \midrule
    \rowcolor{myyellow}
    \multicolumn{5}{l}{\textit{\textbf{Ablation on Data Difficulty}}} \\
    Standard ($\sim$40\% Acc) & 42.7 & 40.6 & 16.7 & \textbf{66.8} \\
    Hard ($\sim$10\% Acc) & 43.0 & 38.6 & 18.4 & 66.7 \\
    Easy ($\sim$90\% Acc) & 48.5 & 0.5 & 51.0 & 48.9 \\
    \bottomrule
    \end{tabular}}
    \captionof{table}{Ablation study on training data difficulty.}
    \label{tab:ablation_difficulty}
      \end{minipage}
\end{figure}

We conduct a sensitivity analysis on the number of sampled responses $G \in \{4, 8, 16\}$ to evaluate its impact on \framename{}. As shown in Table \ref{tab:ablation_g}, a small group size ($G=4$) leads to a significant performance drop, with the Rely score decreasing to 56.3. This is primarily due to the high variance in the solvability indicator $I_{\text{solvable}}$, where limited samples cause the model to frequently underestimate its knowledge boundary by falsely labeling solvable queries, thereby suppressing exploration and resulting in low accuracy (32.2\%). While increasing $G$ to 8 yields the highest Rely score (66.8), further expanding the group size to 16 provides no additional gains in reliability (66.2) but substantially increases the computational overhead of the rollout process. Therefore, we conclude that $G=8$ serves as the optimal configuration, achieving the best balance between the precision of knowledge boundary estimation and overall training efficiency.

\subsection{Ablation on Training Data Difficulty}
\label{app: ablation on Data Difficulty}
To further examine performance under more extreme data distributions, we use the original Llama3.1-8B-Instruct model to sample each training instance 10 times and record the proportion of correct responses and construct two extreme subsets based on the base model's initial performance: a ``Hard'' set (avg. correctness $\sim$10\%) and an ``Easy'' set (avg. correctness $\sim$90\%). As shown in Table \ref{tab:ablation_difficulty}, KARL demonstrates remarkable robustness on the Hard set, achieving a Rely score (66.7) comparable to the Standard setting (66.8). This indicates that our Phase I successfully explores knowledge boundary even from sparse knowledge signals, while Phase II effectively calibrates the remaining uncertain queries into appropriate abstentions.


In contrast, performance significantly degrades on the Easy set, where the model fails to learn abstention ($U=0.5\%$) and suffers from high hallucination rates ($F=51.0\%$). This result stems not from a failure of the reward mechanism, but from a lack of abstention signals in the training data. When nearly all queries are solvable ($I_{\text{solvable}}=1$), the model is never forced to encounter its own limits. These findings highlight that effective abstention alignment is data-dependent: effective abstention learning requires a training distribution with a sufficient ratio of hard queries to provide the necessary negative reinforcement signals for the model to recognize its knowledge boundaries.

\begin{table}[t]
\centering
\small
\renewcommand{\arraystretch}{1.15}
\resizebox{0.55\linewidth}{!}{
\begin{tabular}{lcccc}
    \toprule
    \textbf{Phase I Duration} & \textbf{T ($\uparrow$)} & \textbf{U ($- $)} & \textbf{F ($\downarrow$)} & \color{red!60!black}{\textbf{Rely} ($\uparrow$)}\\
    \midrule
    \rowcolor{myyellow}\multicolumn{5}{l}{\textit{\textbf{Ablation on Transition Timing}}} \\
    0\% & 36.0 & 50.0 & 14.0 & 61.0 \\
    25\% & 40.6 & 40.9 & 18.5 & 64.8 \\
    50\% & 42.7 & 40.6 & 16.7 & \textbf{66.8} \\
    75\% & 43.9 & 32.1 & 24.0 & 65.6 \\
    100\% & 50.6 & 18.0 & 31.4 & 65.4 \\
    \bottomrule
\end{tabular}
}
\caption{
Ablation study on the transition timing from Stage I (Exploration) to Stage II (Abstention Calibration).
}
\label{tab:ablation_timing}
\end{table}

\subsection{Ablation on Transition Timing}
\label{app: ablation on Transition timing}
We investigate the impact of the transition point between Stage I (Exploration) and Stage II (Calibration) by varying the duration of Phase I from 0\% to 100\% of the total training steps. As shown in Table \ref{tab:ablation_timing}, the results exhibit a clear trend: as the duration of Phase I increases, the accuracy ($T$) consistently rises from 36.0\% to 50.6\%, while the abstention rate ($U$) steadily declines from 50.0\% to 18.0\%. This trend validates the distinct functional roles of our two-stage strategy, where Stage I prioritizes driving knowledge recall and Stage II is essential for imposing safety constraints through abstention calibration. Ultimately, the 50\% split (our default configuration) achieves the highest Rely score (66.8), as it strikes the optimal balance between eliciting the model's internal knowledge and maintaining a robust knowledge boundary to suppress hallucinations.

\begin{table}[t]
\centering
\small
\renewcommand{\arraystretch}{1.15}
\setlength{\tabcolsep}{2.5pt} 
\resizebox{1\linewidth}{!}{
\begin{tabular}{lcccccccccccccccc}
\toprule
& \multicolumn{4}{c}{\textbf{NQ}} & \multicolumn{4}{c}{\textbf{TriviaQA}} & \multicolumn{4}{c}{\textbf{BioASQ}} & \multicolumn{4}{c}{\textbf{ARC-C}} \\
\cmidrule(lr){2-5} \cmidrule(lr){6-9} \cmidrule(lr){10-13} \cmidrule(lr){14-17}
\textbf{Setting} & \textbf{T($\uparrow$)} & \textbf{U($-$)} & \textbf{F($\downarrow$)} & \color{red!60!black}{\textbf{Rely}} & \textbf{T($\uparrow$)} & \textbf{U($-$)} & \textbf{F($\downarrow$)} & \color{red!60!black}{\textbf{Rely}} & \textbf{T($\uparrow$)} & \textbf{U($-$)} & \textbf{F($\downarrow$)} & \color{red!60!black}{\textbf{Rely}} & \textbf{T($\uparrow$)} & \textbf{U($-$)} & \textbf{F($\downarrow$)} & \color{red!60!black}{\textbf{Rely}} \\
\midrule
\rowcolor{myyellow}
\multicolumn{17}{l}{\textit{\textbf{Ablation on Strategy}}} \\
\quad w/o Phase II ($\alpha = 50\%$) & 50.6 & 18.0 & 31.4 & 65.4 & 74.9 & 10.6 & 14.5 & 84.4 & 65.6 & 13.8 & 20.6 & 77.5 & 81.5 & 0.8 & 17.8 & 82.2 \\
\midrule
\rowcolor{myyellow}
\multicolumn{17}{l}{\textit{\textbf{Ablation on Mixing Ratio $\alpha$ in Stage I}}} \\
\quad $\alpha = 0\%$ (Pure KAR)  & 36.0 & 50.0 & 14.0 & 61.0 & 62.7 & 29.7 & 7.6 & 83.6 & 52.7 & 39.0 & 8.3 & 76.5 & 79.4 & 6.4 & 14.2 & 85.4 \\
\quad $\alpha = 25\%$  & 42.4 & 35.5 & 22.1 & 65.3 & 68.2 & 22.2 & 9.6 & 85.5 & 61.0 & 22.7 & 16.3 & 78.6 & 82.7 & 2.7 & 14.6 & 85.3 \\
\quad $\alpha = 50\%$ (Default)  & 42.7 & 40.6 & 16.7 & \textbf{66.8} & 67.9 & 23.5 & 8.6 & \textbf{85.9} & 57.7 & 32.8 & 9.5 & \textbf{79.7} & 80.0 & 9.1 & 10.9 & \textbf{88.3} \\
\quad $\alpha = 75\%$  & 39.3 & 47.0 & 13.7 & 64.2 & 62.8 & 31.0 & 6.2 & 84.2 & 53.4 & 37.7 & 8.9 & 76.9 & 76.7 & 13.1 & 10.2 & 88.1 \\
\quad $\alpha = 100\%$ (Binary) & 55.5 & 0.0 & 44.5 & 55.5 & 77.7 & 0.0 & 22.3 & 77.7 & 67.8 & 0.0 & 32.2 & 67.8 & 84.5 & 0.0 & 15.5 & 84.5 \\
\bottomrule
\end{tabular}
}
\caption{
Extended ablation results across all datasets (NQ, TriviaQA, BioASQ, and ARC-C).
}
\label{tab:extended_ablation_all}
\end{table}

\subsection{Generalization of Ablation Results}
\label{appendix: ablation_all_datasets}
To further validate the robustness of our findings, we extend the ablation study of the two-stage training strategy and the mixing ratio $\alpha$ to all out-of-distribution (OOD) datasets. As shown in Table \ref{tab:extended_ablation_all}, the performance trends remain consistent across NQ, TriviaQA, BioASQ, and ARC-C. Specifically, our default configuration ($\alpha=50\%$) consistently achieves the highest \textit{Rely} scores across all evaluated domains. These results confirm that the full KARL framework effectively generalizes the optimal accuracy-abstention trade-off to diverse knowledge intensive tasks and formats.

\begin{figure*}[hb]
    \begin{tcolorbox}[title = {Inference Prompt (Without CoT)}, size=title, colframe = white, colbacktitle = black!65!white]
    \textbf{Input: }You are given a Question. Your task is to answer the question based on factual information in your own knowledge.
    
~\\
    Please adhere to the following guidelines when formulating the answer:
    
    1. If the question contains a false premise or assumption, answer ``invalid question''.
    
    2. If you are uncertain or don't know the answer, answer ``I don't know''.
~\\

    Please directly provide the final answer. The final answer MUST be put in \textbackslash boxed\{\}. For example, \textbackslash boxed\{I don't know\}, \textbackslash boxed\{invalid question\}, \textbackslash boxed\{3 times\}, \textbackslash boxed\{New York\}, etc.

~\\
Question: \{question\} 

~\\
\textbf{Output: } \{answer\}
    \end{tcolorbox}
    \caption{Inference prompt without CoT.}
    \label{fig:prompt_without_CoT_qa}
\end{figure*}

\begin{figure*}[t]
    \begin{tcolorbox}[title = {Inference Prompt (With CoT)}, size=title, colframe = white, colbacktitle = black!65!white]
    \textbf{Input: }You are given a Question. Your task is to answer the question based on factual information in your own knowledge.

~\\
    Please adhere to the following guidelines when formulating the answer:
    
    1. If the question contains a false premise or assumption, answer ``invalid question''.
    
    2. If you are uncertain or don't know the answer, answer ``I don't know''.
~\\

    Please reason step by step and then provide the final answer. The reasoning process must be enclosed within \textless/think\textgreater tags. The final answer MUST be put in \textbackslash boxed\{\}. For example, \textbackslash boxed\{I don't know\}, \textbackslash boxed\{invalid question\}, \textbackslash boxed\{3 times\}, \textbackslash boxed\{New York\}, etc.

~\\
Question: \{question\}

~\\
\textbf{Output: } \{answer\}
    \end{tcolorbox}
    \caption{Inference prompt with CoT.}
    \label{fig:prompt_with_CoT_qa}
\end{figure*}

\begin{figure*}
    \begin{tcolorbox}[title = {Judge Prompt}, size=title, colframe = white, colbacktitle = black!65!white]
    \textbf{Input: }Assume you are a human expert in grading predictions given by a model. You are given a question and a model prediction. Judge if the prediction matches the ground truth answer by following these steps:
~\\

        1: Take it as granted that the Ground Truth is always correct.
        
        2: If the Prediction exactly matches the Ground Truth, return ``t''.
        
        3: If the Prediction does not exactly match the Ground Truth, go through the following steps and likely return ``f''.
        
        4: If the Ground Truth is a number, return ``t'' if and only if the Prediction gives a number that almost exactly matches the ground truth.
        
        5: If the Prediction is self-contradictory, must return ``f''.
        
        6: If the prediction is not answering the question, must return ``f''.
        
        7: If the prediction is a concise and correct summary of the ground truth, return ``t''.
        
        8: If ground truth contains a set of items, prediction must contain exactly same items for the value to return to be ``t''.
        
        9: Otherwise, return ``f''.
~\\

        Question: \{question\}
        
        Ground Truth: \{ground\_truth\}
        
        Prediction: \{solution\_str\}
~\\

        Return only one character: t / f. Do not add any explanation or other text.
    
~\\
\textbf{Output: } \{answer\}
    \end{tcolorbox}
    \caption{LLM-as-a-Judge Prompt.}
    \label{fig:judge_prompt}
\end{figure*}

\end{document}